\documentclass[conference]{IEEEtran}
\IEEEoverridecommandlockouts
\usepackage{cite}
\usepackage{amsmath,amssymb,amsfonts}
\usepackage{algorithmic}
\usepackage{graphicx}
\usepackage{textcomp}
\usepackage{balance}

\usepackage{xcolor}
\def\BibTeX{{\rm B\kern-.05em{\sc i\kern-.025em b}\kern-.08em
    T\kern-.1667em\lower.7ex\hbox{E}\kern-.125emX}}
\begin{document}

\title{Syn2Real Domain Generalization for Underwater Mine-like Object Detection Using Side-Scan Sonar}
\author{Aayush Agrawal $^{1}$, Aniruddh Sikdar$^{2}$, Rajini Makam $^{3*}$, \\Suresh Sundaram$^{3}$, Suresh Kumar Besai$^{4}$ and Mahesh Gopi $^{4}$
\thanks{$^{*}$Corresponding Author}%
   
     \thanks{$^{1}$Aayush Agrawal is with the Department of Chemical Engineering, Indian Institute of Technology, Madras, India
    {\tt\small \{agrawal@smail.iitm.ac.in\}}}%
     \thanks{$^{2}$Aniruddh Sikdar is with Robert Bosch Centre for Cyber-Physical Systems, Indian Institute of Science, Bangalore, India
    {\tt\small \{aniruddhss@iisc.ac.in\}}}%
        \thanks{$^{3}$Rajini Makam and Suresh Sundaram are with the Department of Aerospace Engineering, Indian Institute of Science, Bangalore, India
    {\tt\small \{rajinimakam@iisc.ac.in, vssuresh@iisc.ac.in\}}}%
     \thanks{$^{4}$Suresh Kumar Besia and Mahesh Gopi are with Naval Science and Technology Laboratory (NSTL), Visakhapatnam, India
    {\tt\small \{sureshkumar.b.nstl@gov.in, maheshgopi.nstl@gov.in\}}}%
  }  


\maketitle

\begin{abstract}

Underwater mine detection with deep learning suffers from limitations due to the scarcity of real-world data.
 This scarcity leads to overfitting, where models perform well on training data but poorly on unseen data. This paper proposes a Syn2Real (Synthetic to Real) domain generalization approach using diffusion models to address this challenge. We demonstrate that synthetic data generated with noise by DDPM and DDIM models, even if not perfectly realistic, can effectively augment real-world samples for training. The residual noise in the final sampled images improves the model's ability to generalize to real-world data with inherent noise and high variation. 
The baseline Mask-RCNN model when trained on a combination of synthetic and original training datasets, exhibited approximately a 60\% increase in Average Precision (AP) compared to being trained solely on the original training data. This significant improvement highlights the potential of Syn2Real domain generalization for underwater mine detection tasks.

\end{abstract}


\section{Introduction}


Recent advancements in marine robotics, particularly autonomous underwater vehicles (AUVs) equipped with sophisticated side-scan sonar (SSS) systems, have opened up new possibilities for large-scale ocean exploration tasks like mapping, object detection, and environmental monitoring  \cite{gerg2022, Dinh2023}. While these SSS images provide valuable information for seabed research, manually identifying underwater mines (MLOs) from this data is a time-consuming process \cite{Dura2008}. To address this challenge and improve exploration efficiency, automatic target recognition (ATR) techniques using deep learning are employed. However, underwater mine detection with deep learning faces a unique obstacle: the scarcity of real-world data. Unlike tasks with abundant samples, such as fish detection, underwater mines are rarely encountered, creating a sparse detection problem. This limited data can lead to overfitting, where the model performs well on the specific training SSS images but fails to generalize, potentially missing real mines during actual exploration  \cite{Fu2023, Xu2023S}.
 
Recent studies on MLO detection and segmentation have primarily relied on texture-based, geometric-based, and spectral features, alone or in combination \cite{Fu2023}. RPFNet, a recurrent pyramid frequency feature fusion network, is proposed in \cite{WZZ2023}, addressing detection but not the data scarcity problem. 
In \cite{Gebh2017}, a DNN is utilized for MLO detection in sonar imagery, examining the effects of depth, memory, and training data.

\begin{figure}
    \centering
    \includegraphics[scale=0.2]{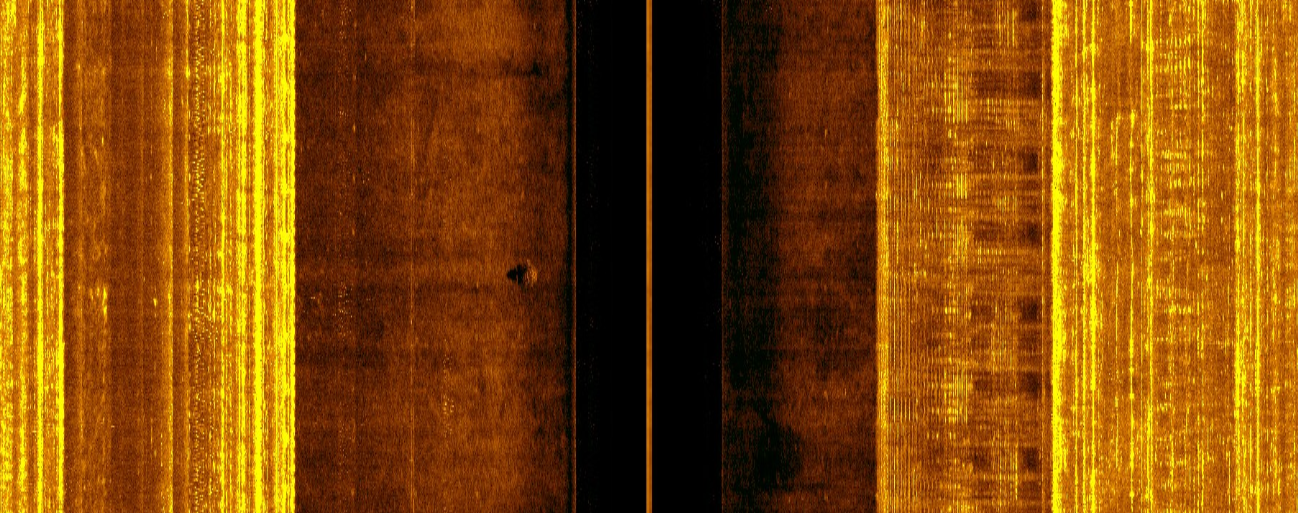}
    \caption{\textcolor{blue}{}}
    \label{fig:Arch}
\end{figure}
 

Two widely explored strategies for addressing data scarcity are zero-shot learning and synthetic data generation. Zero-shot learning allows models to classify new instances without specific training on them \cite{Socher2013}. In \cite{Xu2023}, Xu et al. proposed MFSANet, a method using deep neural networks (DNNs) to address zero-shot learning challenges. MFSANet utilizes optical-acoustic image pairs to create pseudo-SSS images. 
Domain generalization aims to train models on synthetic data and perform inference on real-world images by minimizing the domain gap between the two \cite{choi2021robustnet, Chattopadhyay_2023_ICCV}. Synthetic data generation addresses the scarcity of data by creating artificial data that mirrors the properties of real data \cite{Goodfellow2014}, thereby helping to achieve the goal of domain generalization\cite{udupa2023mrfp}.
Li et al. \cite{Li2021} used style transfer to transform an optical target image into a background image from side-scan sonar, resulting in a simulated image. This approach yielded a target classification accuracy exceeding 75\%.  

In addressing data scarcity in side-scan sonar applications, various approaches have been explored. Initially, efforts to augment data through synthetic generation methods were hindered by limited efficacy, leading to continued overfitting of deep learning models \cite{b1}. To overcome this, a GAN \cite{Bore2020} was developed, with the aim of enhancing generalization capabilities by generating synthetic sonar data, albeit at the cost of significant computational resources.
Further advances were made in \cite{Yang2024}, where a few-shot underwater object augmentation method was proposed for multitask scenarios. Leveraging diffusion models and transfer learning, they fine-tuned optical pre-trained models for SSS samples, achieving a promising mIoU of 74.08\%.

In this paper, we investigate various approaches for generating synthetic images to create a dataset tailored for underwater perception, specifically focusing on Mine-Like Objects (MLOs) for domain generalization. Our contributions are multifaceted. First, while diffusion models are widely known, we present their novel application for generating SSS images of MLOs, with significant improvements achieved through hyperparameter fine-tuning. Additionally, we adapt the Mask RCNN framework to better handle the unique characteristics of synthetic SSS data, enhancing the effectiveness of semantic modeling in mine detection tasks. We also provide a comparative analysis of different synthetic data generation techniques and their performance with Mask RCNN for instance segmentation as a downstream task. To address the lack of available datasets, we generate a custom SSS dataset featuring MLOs, and our study highlights the effective synthetic data models for domain generalization in SSS imagery, improving model performance with limited annotated data.


The structure of the paper is as follows: Section \ref{pf} gives an outline of the problem statement and the dataset used. Section \ref{s2} presents an overview of the synthetic data generation techniques employed for optical and SSS images. In Section \ref{s3}, the evaluation and results obtained with GAN and diffusion models are presented. Concluding remarks are provided in Section \ref{s5}.

\begin{figure*}
    \centering
    \includegraphics[scale=0.5]{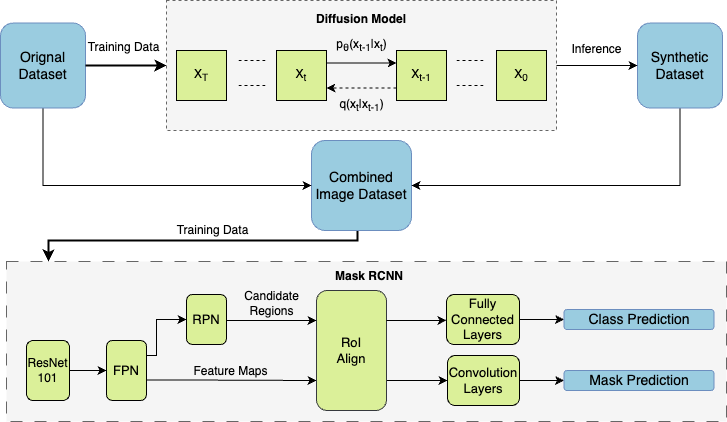}
    \caption{\textcolor{blue}{}}
    \label{fig:Arch}
\end{figure*}

\section{Problem Formulation}
\label{pf}

The purpose of this study is 2 fold. The first goal is to modify and tune synthetic data generation models for side scan sonar images, the second goal is to evaluate the synthetic data generation models on their ability to generalize the domain from synthetic to real using a segmentation model. \\
A dataset of side scan sonar images was collected for 2 types of mines, conical and cylindrical as explained in \ref{sec:datasets}. These images (Original Side Scan Sonar Images) were used to train synthetic data generation models. Additionally, a dataset comprising optical underwater images was also used to train the synthetic data generation models, this was done solely to compare the data sources for the task at hand. Using all the synthetic and original data, multiple datasets were generated by combining data from different sources, and then a semantic segmentation model was trained on these datasets, and evaluated.

\subsection{Datasets}
\label{sec:datasets}
Although SSS images are the main focus of this study, we also experimented with synthetic images generated using optical-underwater datasets. We used the Semantic Segmentation of Underwater Imagery (SUIM) dataset \cite{islam2020semantic} for optical images. It consists of over 1500 pixel-level annotated images taken with an optical camera that cover eight different object categories: the sea floor, human divers, invertebrate reefs, aquatic plants, wrecks/ruins, and vertebrate fish. 1906 $\times$ 1080 is the resolution of the image.

Due to the lack of open-source SSS data, we created a novel dataset consisting of MLOs. The data was collected using the Starfish454 OEM with 450KHz frequency fitted on ROV and Sea Scan ARC Scout MkII operating at 600 KHz and 1200 KHz fitted on AUV. The side scan sonar images are gathered by deploying cylindrical and truncated conical-shaped mine-like objects in captive waters and harbors. The images are captured at different dynamic conditions such as varied depth, distance, and orientation. A total of 461 images were captured, consisting of 269 cylindrical mines and 192 conical mines.

\section{Methodology} \label{s2}
This section first discusses the synthetic data generation models and the hyperparameters used. Further, the semantic segmentation model and the training data used have been explained. An overview of the architecture of the diffusion model with Mask RCNN can be found in Fig. \ref{fig:Arch}. The green boxes represent the trainable segments of the workflow, and the blue boxes represent the datasets used for training. "Original Dataset" refers to the true SSS image dataset. "Synthetic Dataset" represents The images generated via the Diffusion Model.

The synthetic data generation models compared are, DCGAN and diffusion models with different noise schedulers(DDPM and DDIM).
DCGAN \cite{radford2015unsupervised} is a model built primarily for picture creation. It uses deep convolutional neural networks in both the generator and discriminator. In our experiments, we use the conventional DCGAN loss function as given in \cite{radford2015unsupervised}.

Diffusion models \cite{b5} rival GANs in generating high-quality images by reconstructing data from an initial distribution.  At their core lies the captivating interplay between a forward noising process and a reverse denoising process\cite{dhariwal2021diffusion}. In \cite{SWM2015} an algorithm was introduced to model probability distributions, which allows exact sampling and probability evaluation. This algorithm focused on estimating the reversal of a Markov diffusion chain, mapping data to a noise distribution.

\begin{eqnarray}   
q(x_{t}|x_{t-1})=\mathcal{N}(x_{t};\sqrt{1-\beta_{t}}x_{t-1},\beta_{t}I) \label{eq1dm} \\q(x_{1:T}|x_{0})=\prod_{t=1}^{T}q(x_{t}|x_{t-1})\label{eq2dm}
\end{eqnarray}
where $x_{t}$ is the image state. The term $\beta$ is known as the “diffusion rate” and is precalculated using a “variance scheduler”. 
Using reparameterization, $\alpha_{t}=1-\beta_{t}, \bar{\alpha}_{t}=\prod_{t=1}^{T}\alpha_{t}$
\begin{eqnarray}    
x_{t} &=&\sqrt{\alpha_{t}}x_{t-1}+\sqrt{1-\alpha_{t}} \epsilon_{t-1}\nonumber \\ & =&\sqrt{\alpha_{t}\alpha_{t-1}}x_{t-2} + \sqrt{1-\alpha_{t}\alpha_{t-1}} \bar{\epsilon}_{t-2} \nonumber\\
& = & \sqrt{\bar{\alpha}_{t}}x_{0}+\sqrt{1-\bar{\alpha}_{t}}\epsilon  \label{eq3dm}
\end{eqnarray} 

where the image state \(x_t\) at time \(t\) is influenced by its past values and random noise terms. It is expressed as a combination of its previous value \(x_{t-1}\) weighted by \(\sqrt{\alpha_{t}}\) and a noise term \(\sqrt{1-\alpha_{t}} \epsilon_{t-1}\), where \(\alpha_{t}\) represents the weight. This relationship extends to the value at time \(t-2\), incorporating the cumulative effect of \(\alpha_{t}\) and \(\alpha_{t-1}\). The evolution of \(x_t\) from an initial condition \(x_0\) is determined by \(\bar{\alpha}_{t}\), representing the cumulative product of \(\alpha_{t}\). These expressions capture the dynamic nature of \(x_t\) over time, influenced by its past values and random fluctuations as given in eq. \ref{eq3dm}. 


The reverse denoising process in diffusion models involves removing noise from a degraded image to reconstruct the original version. In the reverse denoising operation, the model starts with a noisy image and iteratively applies transformations to reduce the noise, with the aim of approximating the initial image. This is facilitated by a U-net for effective noise removal and feature preservation.



There are two popular diffusion model methodologies in literature: Denoising Diffusion Probabilistic Models (DDPM) \cite{dhariwal2021diffusion} and Denoising Diffusion Implicit Models (DDIM) \cite{song2020denoising}. DDPM excels in generating high-quality images without adversarial training but is computationally expensive due to its Markov chain simulation, requiring 1000 steps for high-quality results. To address this, DDIM introduces non-Markovian diffusion, allowing faster sampling (50–100 steps) while maintaining competitive image quality.

However, this efficiency comes at the cost of stability and flexibility, as DDIM requires parameter tuning and may result in lower image quality compared to DDPM. In contrast, DDPM ensures better image quality and stability, albeit with slower sampling.

After several rounds of hyperparameter tuning stable results were obtained with the following hyperparameters: For training 100 images were used from the Original dataset from each mine type.
The diffusion models (DDPM and DDIM) are trained for 200 epochs. The beta start and beta end were set to 0.0001 and 0.02 respectively. The beta scheduler was set to a linear model. The dcGAN model is trained for 50 epochs (until convergence). The batch size for training all the models was 16.

Once the datasets have been created using the synthetic models, a semantic segmentation model namely a Mask RCNN \cite{he2017mask} is trained on these. A Mask RCNN features a backbone network for feature extraction, a Region Proposal Network (RPN) for bounding box proposals, and a Mask Head for pixel-level segmentation. The backbone captures hierarchical features, while the RPN generates candidate object proposals.

Only conical mines and Side Scan Sonar images were utilized for this experiment. The DCGAN synthetic data was not used further due to reasons explained in \ref{sec:results}. A total of 600 images of which 200 generated images from DDIM and DDPM models and 200 images from the Orignal dataset (with augmentations) were annotated and subsequently used to train a semantic model. All annotations were provided manually. Seven datasets were created using combinations of DDIM, DDPM, and Original Images. Namely, Original, DDPM, DDIM, DDPM+DDIM, DDPM+Original, DDIM+Original, DDPM+DDIM+Original. For verification, a new set of 100 original conical mine images was used. The SSS semantic segmentation model was trained for over 50 epochs with a batch size of 4 using Binary Cross-Entropy + Focal (BCE + Focal). The initial model weights were set from a pre-trained model on the MS-COCO Dataset.

\section{Results and Discussion}  \label{s3}

In this section, detailed qualitative and quantitative comparisons of the generative methods, GAN \& diffusion models (DDPM and DDIM) are presented for both optical and SSS images. The quality of the generated images is evaluated using heuristic metrics like  Fréchet Inception Distance (FID), Kernel Inception Distance (KID), Inception Scores (IS), and Object Reconstruction Rate (ORR).
Further, the qualitative assessment of DDPM and DDIM are presented. This is followed by the evaluation of their performance for domain generalization for semantic segmentation. The metrics used are Average Precision (AP) with different IOU and Area under Precision vs IOU Threshold Curve (AUPC) scores.

\begin{table*}
\caption{Comparison of Generative Methods For Synthetic Data}
{\renewcommand{\arraystretch}{1.3}%
\begin{center}
\begin{tabular}{|c | c | c | c | c| c | c|c|c|c|}
\hline 
\textbf{Metric} &\multicolumn{9}{|c|}{\textbf{Method}} \\
\cline{2-10} 
&\multicolumn{3}{|c|}{\textbf{\textit{Conical Mines (SSS images)}}} & \multicolumn{3}{|c|}{\textbf{\textit{Cylinderical Mines (SSS images)}}} &\multicolumn{3}{|c|}{\textbf{\textit{Optical Images}}}
\\
\cline{2-10} 
 & DDPM & DDIM & dcGAN & DDPM & DDIM & dcGAN & DDPM & DDIM & dcGAN\\
\hline
FID &  135.71  & 251.89 & 292.3 & 171.02 & 231.41 & 303.76 & 197.03 & 231.67 & 302.34\\
\hline
KID & 0.105 & 0.293 & 0.361 & 0.082 &  0.252 & 0.356
 & 0.089 & 0.263 & 0.312 \\
\hline
IT  & 159.35  & 15.10 & 3.50 & 161.02 & 14.70 & 3.47  & 160.12  & 15.60 & 3.76 \\
\hline
ORR   & 0.78 & 0.64 & 0.34 & 0.68 & 0.60 & 0.26 & - & - & -\\
\hline
\end{tabular}%
\label{tab3:nums}
\end{center}}
\end{table*}
\subsection{Evaluation metrics}
As previously indicated, we employ heuristic criteria to assess the performance of generative models.
The FID is calculated as given in eq. \ref{eq1dm:fid}:
\begin{equation}
\label{eq1dm:fid}
   FID =  \left\| \mu_{\text{real}} - \mu_{\text{fake}} \right\|^2 + \text{Tr} \left( \Sigma_{\text{real}} + \Sigma_{\text{fake}} - 2 \left( \Sigma_{\text{real}} \Sigma_{\text{fake}} \right)^{1/2} \right)
\end{equation}
where $\mu$ shows the mean of the feature vector and $\Sigma$ shows the covariance matrix of the image.
The KID is calculation is provided in eq. \ref{eq1dm:kid}:
\begin{eqnarray}
\label{eq1dm:kid}
    KID & = &     \frac{1}{n^2} \sum_{i=1}^{n} \sum_{j=1}^{n} k(x_i, x_j) - \frac{2}{n^2} \sum_{i=1}^{n} \sum_{j=1}^{m} k(x_i, y_j) \nonumber \ \\ & + & \frac{1}{m^2} \sum_{i=1}^{m} \sum_{j=1}^{m} k(y_i, y_j)
\end{eqnarray}
where $x_{i}$ and $y_{i}$ are real(empirical) feature representations and general feature representations respectively. The choice of the kernel function \(k\) can vary. However, a commonly used one is the Gaussian (RBF) kernel as given in equation \ref{eq:gauss}.
\begin{equation}
\label{eq:gauss}
    k(x, y) = \exp\left(-\frac{\|x - y\|^2}{2\sigma^2}\right)
\end{equation} 

For ORR, we qualitatively assess image quality by counting distinctive mine images with clear boundaries and shadows.



To evaluate the amount of noise in the image, and how it affects the semantic segmentation model we calculate the standard deviation of the pixels in the image and the SNR value. Then, the Signal-to-Noise Ratio (SNR) is given by:
\begin{equation}
    \text{SNR} = 10 \cdot \log_{10}\left(\frac{P_s}{P_n}\right) 
\end{equation}

where $P_s$ represents the signal calculated by taking a mean of all pixel values and $P_n$ represents the noise calculated by taking the standard deviation of all pixel values. This formula is often expressed in decibels (dB) to provide a more human-readable scale for the SNR value.

Finally, semantic models trained on labeled data are tested for downstream analysis. The performance is analyzed using Average Precision (AP) with different IOU and Area under Precision vs IOU Threshold Curve (AUPC) scores. The formulae for which are given in equation \ref{eq1dm:ap}. TP refers to true positive and FP refers to false positive.

\begin{equation}
\label{eq1dm:ap}
\text{{AP}}_k = \frac{\text{{TP at IoU}} \geq k}{\text{{TP at IoU}} \geq k + \text{{FP at IoU}} \geq k}.
\end{equation}

\subsection{Results} \label{sec:results}




We first compared the performance of models applied to optical images versus side-scan sonar (SSS) images. Due to the lack of publicly available optical datasets with Mine-Like Objects (MLOs), our focus was on evaluating the models’ ability to generate synthetic data and their performance with these datasets. Our quantitative analysis showed that models trained on optical images performed worse than those trained on SSS images. Qualitative results also revealed that optical images, with their tendency to capture excessive background features, often obscured key details of MLOs, making it harder to distinguish and represent them clearly. \\
Optical images, while convenient for synthetic image generation, exhibited higher FID scores compared to SSS images, indicating less accurate reconstructions. This issue likely arises from the poor clarity of optical images in challenging environments with low light, dust, and water interference, which SSS images better handle. The superior performance of SSS images reinforces their suitability for MLO detection.

From the results in Table \ref{tab3:nums}, it is evident that DDPM performs better than DDIM in terms of data generation. This is evident from the lower FID and KID scores. However, we can see an almost 10x difference in the time inference time, with DDIM performing better. dcGAN has consistently performed poorly for all 3 metrics used but has the lowest inference time.

\begin{figure}
    \centering
    \begin{tabular}{ccc}
      \includegraphics[width=1 in]{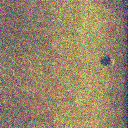}   & \includegraphics[width=1 in]{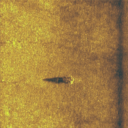}  &      
     \includegraphics[width=1in]{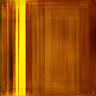} \\
      (a) & (b) & (c)\\
       \includegraphics[width=1in]{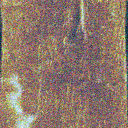} &
    \includegraphics[width=1in]{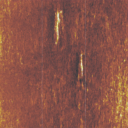} &
    \textbf{\includegraphics[width=1in]{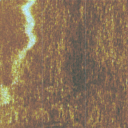}}
    \\
    (d) & (e) & (f)
    \end{tabular} 
    \caption{Synthetically generated images of $M_1$ (Conical mines): (a)~DDIM (b)~DDPM (c) GAN and $M_2$ (Cylindrical mines):~(d)~DDIM (e)~DDPM (f) GAN}
    \label{fig:m1}
\end{figure}

A visual assessment of the generated images is conducted; some examples of the sampled images are shown in Fig. \ref{fig:m1}. Specifically, DDPM exhibits superior capability in rendering clear representations of mines and their associated shadows, as evident from the ORR score. In particular, the images generated by DDIM display a noticeable amount of noise, as observed in Fig. \ref{fig:m1} (a) and (d). In contrast, DDPM-generated images illustrated in Fig. \ref{fig:m1} (b) and (e), present well-defined boundaries and minimal noise. The results shown in Table \ref{tab3:noise_snr} support these conclusions. Furthermore, images produced by the DCGAN model show reconstructed backgrounds accurately, as depicted in Fig. \ref{fig:m1} (c) and (f). The ORR values in Table \ref{tab3:nums}, further suggest that these synthetic images struggle to capture crucial details like mines and shadows. Hence, the DCGAN-generated images are not considered for the segmentation task due to their incredibly low ORR.

\begin{figure*}
    \centering
    \includegraphics[scale=0.28]{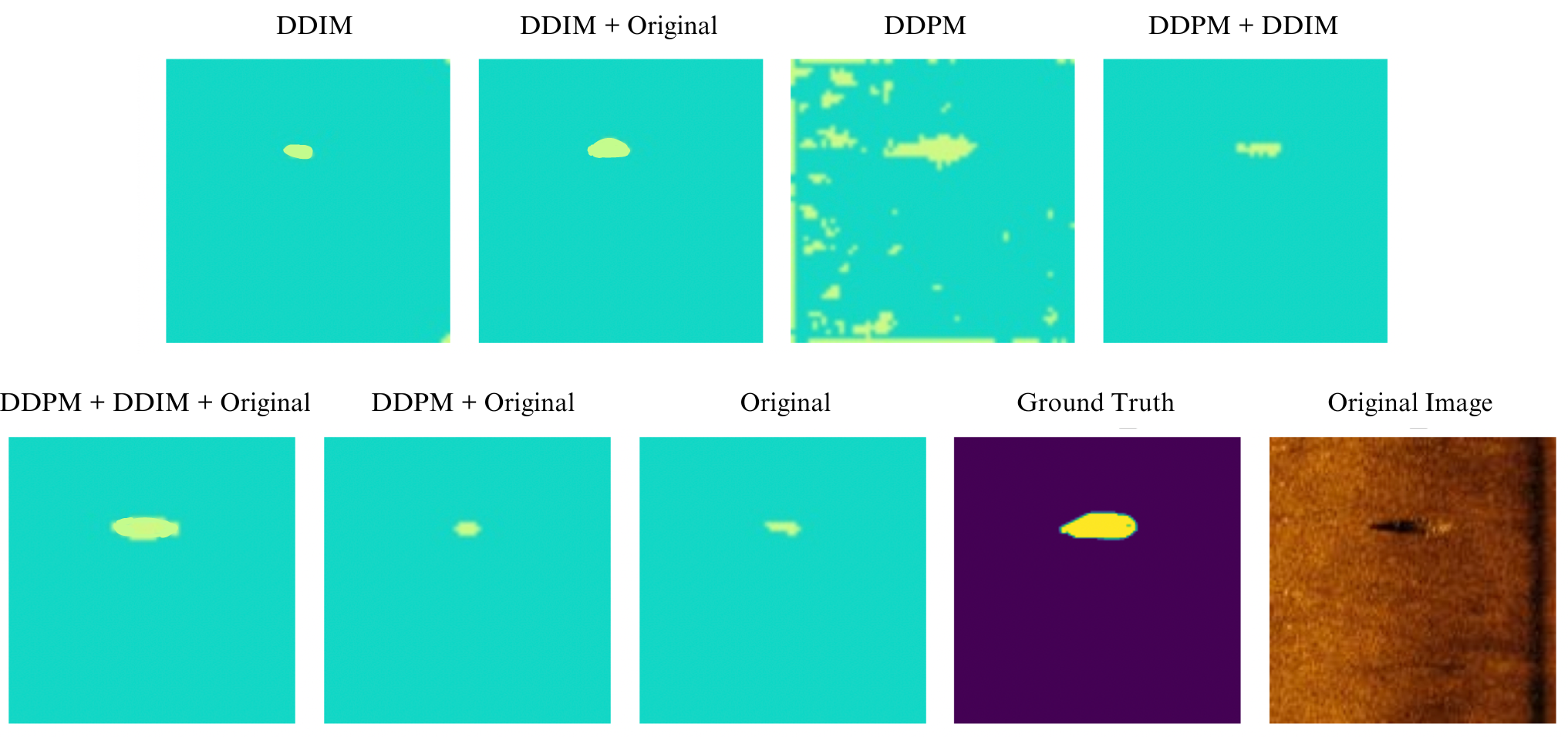}
    \caption{Segmentation Results from Model Trained on Different Datasets}
    \label{fig:Arch}
\end{figure*}

\begin{table}
  \centering
{\renewcommand{\arraystretch}{1.3}%
  \caption{Standard Deviation and SNR for Side Scan Sonar Images}
  \label{tab3:noise_snr}
  \begin{tabular}{|c|c|c|}
    \hline
    \textbf{Model} & \textbf{Average Noise} & \textbf{Average SNR (dB)} \\
    \hline
    DDPM M1 & 23.285 & 7.170 \\
    \hline
    DDPM M2 & 22.489 & 7.363 \\
    \hline
    DDIM M1 & 24.768 & 6.998 \\
    \hline
    DDIM M2 & 24.768 & 7.132 \\
    \hline
  \end{tabular}}
\end{table}
We also assess the robustness of diffusion models to noise in terms of performance. When compared to DDPM-generated images, the DDIM-generated images exhibit higher levels of noise. Table \ref{tab3:noise_snr} illustrates this; in comparison to DDPM, DDIM has roughly 10\% higher noise and a 3\% decrease in SNR. 

\begin{table*}
{\renewcommand{\arraystretch}{1.1}
\caption{Precision Values for Model Trained on Different Datasets}
\begin{center}
\begin{tabular}{|c|c|c|c|c|c|c|}
\hline 
Training Dataset & $AP_{50\%}$ & $AP_{75\%}$ & $AP_{90\%}$ & $AP_{50:95\%}$ & $Avg_{50\%,75\%,90\%}$ & AUPC \\
\hline
Orignal & 0.682 & 0.417 & 0.0 & 0.399 & 0.366 & 0.168 \\
\hline
DDPM & 0.434 & 0.212 & 0.0 & 0.263 & 0.215 & 0.096 \\
\hline
DDIM & 0.727 & 0.633 & 0.0 &0.521 & 0.442 & 0.217 \\
\hline
DDIM+Orignal & 0.808 & 0.722 & 0.167 & 0.602 & 0.566 & 0.257 \\
\hline
DDPM+Orignal & 0.818 & 0.622 & 0.0 & 0.538 & 0.473 & 0.226 \\
\hline
DDPM+DDIM & 0.759 & 0.667 & 0.0 &  0.551 & 0.475 & 0.228 \\
\hline
DDPM+DDIM+Orignal & 0.833 & 0.737 & 0.167 & 0.635 & 0.579 & 0.264 \\
\hline
\end{tabular}
\label{tab3:sem}
\end{center}}
\end{table*} 

The findings from the semantic model yield intriguing outcomes as seen in Table \ref{tab3:sem}, as they reveal a reversal in performance trends. Although DDPM demonstrates significantly better image generation quality, DDIM-generated image datasets outperform DDPM for domain generalization from synthetic to real. Observing the results, it is evident that combining all three datasets yields the most favorable outcomes. However, across all metrics used, when the model is trained on datasets consisting or solely of DDIM it consistently outperforms their counterpart datasets based on DDPM. We see an almost 2x increase in performance when trained on just synthetic images and when trained on a combination of Original + Synthetic, we see an increase of roughly 14\%. The best results were noted to be from DDPM+DDIM+Original with an AUPC of 0.264 which is closely followed by DDIM+Original with an AUPC of 0.257. This indicates that adding the DDPM images to the dataset didn't have as much of an impact. The lowest was for only DDPM, 0.096.


The DDIM model's infusion of noise into the sampled images is partially responsible for the observed variability in the results. Previous studies, including \cite{Zhang2021, noh2017regularizing}, have shown that adding noise and increasing variation can enhance the generalization of semantic segmentation models, even though this noise may result in higher FID and KID scores. Results indicate that noise addition blocks overfitting caused by regularization (\cite{noh2017regularizing}).


\section{Conclusion} \label{s5}

In this paper, we examined methods for generating SSS images specifically for Syn2Real (synthetic to real) domain generalization. We aim to bridge the gap between training on synthetic data and achieving good performance on real-world SSS data.

We investigated two generative models: GANs and diffusion.  While both were explored for optical and SSS images, limitations in GAN-generated and optical images led us to focus on SSS data and diffusion models. 
We achieved significant performance improvements on downstream tasks. Combining synthetic data with the original dataset led to impressive AP and AUPC scores of 83.3\% and 0.264, respectively, using Mask R-CNN for semantic segmentation. By using synthetic images along with original images for training we can see an increase in performance of about 60\% from the model trained on just the original images. This shows that the diffusion model can generate varied synthetic data that can significantly improve domain generalization and the overall performance of underwater mine detection. 

Our analysis revealed that the DDIM sampling method produced the most effective synthetic SSS images for domain generalization. This is because the final sampled images from DDIM contains a higher degree of noise, leading to more diverse and generalizable datasets.


\balance

\bibliographystyle{ieeetr}
\bibliography{refe}
\end{document}